\documentclass{ecai}
\usepackage{times}
\usepackage{graphicx}
\usepackage{latexsym}
\usepackage[dvipsnames]{xcolor}
\usepackage{soul}
\usepackage{siunitx}
\usepackage{xcolor}
\usepackage{tikz}
\usetikzlibrary{calc}
\usepackage{amsmath}
\usepackage{caption}
\usepackage{wrapfig}
\usepackage{booktabs}
\usepackage{multirow}
\usepackage{url}
\definecolor{plot_blue}{RGB}{31,119,180}
    
\newcommand{\rhob}{$\rho_b$} 

\newcommand{\vp}{$V_p$} 
\newcommand{\gr}{$\gamma$} 
\newcommand{\nphi}{$n_\phi$} 
\newcommand{\las}{\texttt{LAS}} 
\newcommand{\nlog}{\textsc{nlog}}

\begin{document}

\title{Machine Learning for Gas and Oil Exploration}

\author{Vito Alexander Nordloh\institute{ICSA, The University of Edinburgh, email: v.a.nordloh@sms.ed.ac.uk} %
\and Anna Roub\'i\v{c}kov\'a\institute{EPCC, The University of Edinburgh, email: a.roubickova@epcc.ed.ac.uk} 
\and Nick Brown\institute{EPCC, The University of Edinburgh, email: n.brown@epcc.ed.ac.uk} }

\maketitle
\bibliographystyle{ecai}

\begin{abstract}
Drilling boreholes for gas and oil extraction is an expensive process and profitability strongly depends on characteristics of the subsurface. As profitability is a key success factor, companies in the industry utilise well logs to explore the subsurface beforehand. These well logs contain various characteristics of the rock around the borehole, which allow petrophysicists to determine the expected amount of contained hydrocarbon. However, these logs are often incomplete and, as a consequence, the subsequent analyses cannot exploit the full potential of the well logs.

In this paper we demonstrate that Machine Learning can be applied to \emph{fill in the gaps} and estimate missing values. We investigate how the amount of training data influences the accuracy of prediction and how to best design regression models (Gradient Boosting and neural network) to obtain optimal results. We then explore the models' predictions both quantitatively, tracking the prediction error, and qualitatively, capturing the evolution of the measured and predicted values for a given property with depth. Combining the findings has enabled us to develop a predictive model that completes the well logs, increasing their quality and potential commercial value. 
\end{abstract}

\section{Introduction}%
\label{sec:intro}

\noindent Increasing computation power and recent advances in Machine Learning (ML) give rise to numerous applications of ML techniques to real world problems. The petroleum industry, with its wealth of data, is well positioned to leveraging these approaches to deliver significant benefit. This available data typically consists of measurements of physical properties gathered while drilling exploratory boreholes and recorded in \emph{well logs}, which are sequential records of properties recorded at regular depth increments.

The well logs serve as input for various modelling tools which are used to infer properties of the geology surrounding the well. Ultimately, these aid commercial decision making around further development of the well or even entire oilfields, based on the estimated presence of hydrocarbons. The process of well log interpretation, which converts the raw measurements into commercially valuable information, requires significant human effort and great expertise and experience, making the process both time consuming and expensive. Furthermore, inaccurate prediction of the hydrocarbon contents has a significant economical impact, either due to missed opportunities or due to the costs associated with drilling a production well with low hydrocarbon yield. 

The well logs record \emph{raw} measurements collected by various tools, and are naturally noisy and incomplete. Petrophysicists tasked with the interpretation therefore spend a non-trivial amount of effort on conditioning, otherwise known as cleaning, the well logs by adjusting wrong measurements and estimating the missing values. Only once this has been completed can they proceed with the interpretation of the rock properties. 

A recent study \cite{swoop_cug} attempted to bring the time needed for a well log interpretation down from 7 days of human effort to 7 minutes of automated computation with minimum human input using ML. One of the main obstacles identified in that work was the amount of data missing in the logs. This is because the measurements are used as an input for the ML models and their occasional absence severely limits either the volume of usable data or the choice of applicable models.
%
%
The benefit of the work described in this paper is twofold, if one can estimate values of the well log gaps accurately enough it will enable petrophysicists to derive interpretations with greater certainty, and also unlock a multitude of additional ML techniques for further processing of the well logs.
%

In this work we use supervised ML techniques to derive the relationships among different physical properties measured within a borehole.  In turn, we use one of these properties as a target (or dependent) variable, while the other properties serve as input (or independent) variables. Every record in a well log represents one observation, and we train the models using a subset of these observation, maintaining a disjoint set for evaluation. For both training and evaluation purposes we only use data where the target property has been recorded, using the measured value as a ground truth. This ultimately avoids  the need for a petrophysicist to label the data beforehand, and it is our hypothesis that this will result in more consistent data values, avoiding human introduced biases. 


The paper is structured as follows, Section ~\ref{sec:context} provides context to this work, describing background information and a related work overview.
In section~\ref{sec:method} we report on two different ML techniques used to predict the missing values, together with their accuracy, presenting the most suitable configuration. Section~\ref{sec:alternative_approaches} explores the impact of using larger training data set, and we summarise our findings in Section~\ref{sec:conclusion}.


\section{Background and Related Work}%
\label{sec:context}

This section briefly introduces well logging and the applied ML techniques.
Prior work concerned with missing data in well logs is presented at the end of this section.

\subsection{Well Logging}
\begin{figure}
    \centering
    \includegraphics[width=.9\linewidth]{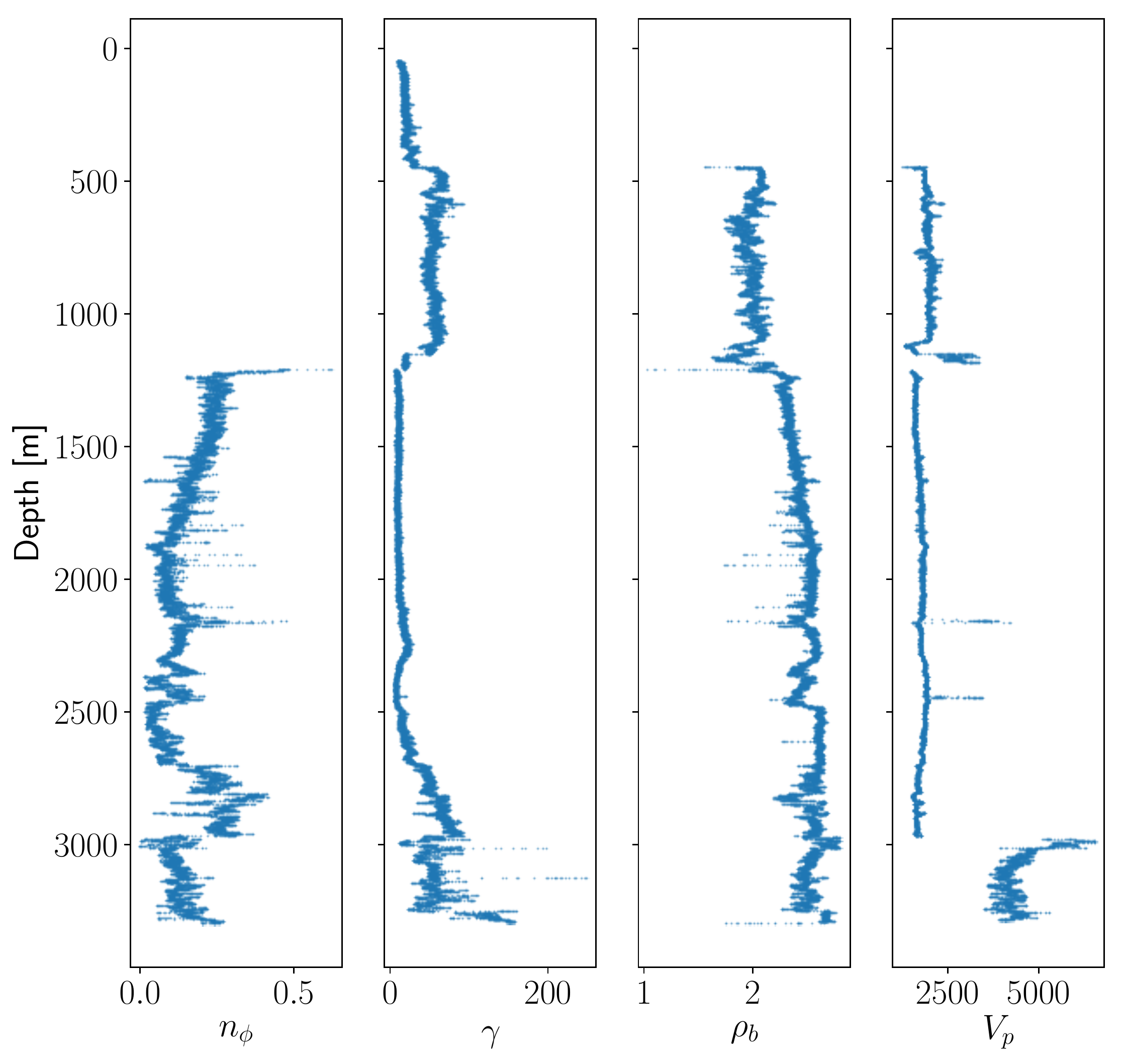}
    \caption{Plot of well L17-02 with its four properties \nphi{}, \gr{}, \rhob{} and \vp{}. Contrary to usual graphs, the $y$-axis is reversed to intuitively represent the borehole's depth.}
    \label{fig:well_log}
\end{figure}

In the gas and oil industry, \emph{well logging}~\cite{crain2002crain, 2007WLfE} is the process of lowering measuring devices, \emph{sondes}, down a borehole and simultaneously collecting various physical characteristics of the traversed rock formations at regular depth steps. This raw data is manually analysed by petrophysicists who determine, amongst other properties, expected hydrocarbon yield. Recorded values of these properties are typically plotted on a vertical graph next to each other, with depth on the $y$-axis and the property on the $x$-axis, and an example of this is illustrated in Figure~\ref{fig:well_log}. This provides an intuitive representation of the borehole and characteristics of different formations, enabling petrophysicists to draw inference about the mineralogy and presence of hydrocarbons at specific depths. In order to utilise most of the available well logs, we restrict this work to the most commonly measured rock properties which are neutron porosity, gamma ray, bulk density, and sonic.

\noindent \emph{Neutron Porosity} (\nphi{}) is a unitless measure corresponding to the porosity of the rock and the fluid present in the pores, making it an indirect indicator of the contained hydrocarbon volume.
%
The \emph{Gamma Ray} (\gr{}) value is a measurement of the natural gamma radiation emitted by surrounding rocks, and is used to derive information about mineralogical composition of the surrounding geology. 
\emph{Bulk Density} (\rhob{}) represents the consistency of the rock and is derived from the rock's absorption of artificially emitted gamma rays. 
The porosity and lithology are further informed by \emph{Sonic} measurements (\vp{}), which record the time required for an acoustic signal to travel a specific distance through a rock formation. This is informative because it is affected by both the rock's porosity and the type of fluid present in its pores. 

The values of these properties are measured and recorded at regular depth intervals, but the challenge is that some values might be missing at specific levels. There are various reasons for missing data, for instance a malfunction of the sondes or human error. Even monetary pressures can drive the omission at specific depths, where some measurements are considered too expensive to be cost effective to gather. 

\subsection{Machine Learning}
Supervised ML automatically derives a functional relationship between input and output variables based on a set of existing training data. Every input is an ordered set (a vector) of values, known as independent variables (or features), which describes various properties of the problem. The output values, known as dependent, or target, variables, are predicted by evaluating the learned function over the input vector. 
Depending on the domain of the target variables, one distinguishes between \emph{classification}, where the target values form a (usually relatively small) finite set, and \emph{regression}, for targets with a continuous, potentially infinite domain.

In order to approximate the missing data within the logs we applied two regression models: neural networks and Gradient Tree Boosting. Although both models accept a vector of features as input and output a target value, they are fundamentally different.

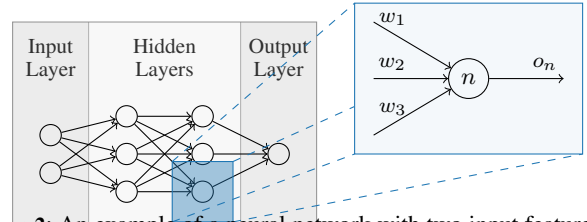
\begin{figure}
    \centering
    \begin{tikzpicture}[
        neuron/.style={circle,draw,inner sep=0,minimum size=8pt},
        bigneuron/.style={circle,draw,inner sep=0,minimum size=15pt}
    ]
        \fill[gray,opacity=0.15] (-0.5,-1) rectangle ++(1,2.75);
        \fill[gray,opacity=0.055] (0.5,-1) rectangle ++(2,2.75);
        \fill[gray,opacity=0.15] (2.5,-1) rectangle ++(1,2.75);
        
        \draw[gray,opacity=0.5] (0.5,-1) -- ++(0,2.75);
        \draw[gray,opacity=0.5] (2.5,-1) -- ++(0,2.75);
        
        \draw[gray,opacity=0.8] (-0.5,-1) rectangle ++(4,2.75);
        
        \draw (0, 1.25) node[opacity=0.8,font=\footnotesize,align=center] {Input\\Layer};
        \draw (1.5, 1.25) node[opacity=0.8,font=\footnotesize,align=center] {Hidden\\Layers};
        \draw (3, 1.25) node[opacity=0.8,font=\footnotesize,align=center] {Output\\Layer};
        
        \draw (0,0.25) node[neuron] (i1) {};
        \draw (0,-0.25) node[neuron] (i2) {};
        
        \draw (1,0.5) node[neuron] (h11) {};
        \draw (1,-0.5) node[neuron] (h12) {};
        \draw (1,0) node[neuron] (h13) {};
        
        \draw (2,0.5) node[neuron] (h21) {};
        \draw (2,-0.5) node[neuron] (h22) {};
        \draw (2,0) node[neuron] (h23) {};
        
        \draw (3,0) node[neuron] (o1) {};
        
        \foreach \i in {1,2}
            \foreach \j in {1,2,3}
                \draw[->] (i\i) -- (h1\j);
                
        \foreach \i in {1,2,3}
            \foreach \j in {1,2,3}
                \draw[->] (h1\i) -- (h2\j);
                
        \foreach \i in {1,2,3}
                \draw[->] (h2\i) -- (o1);
                
        \draw[plot_blue] ($(h22)-(0.4,0.4)$) rectangle ++(0.8,0.8);
        \fill[plot_blue,opacity=0.4] ($(h22)-(0.4,0.4)$) rectangle ++(0.8,0.8);
                
        \draw[plot_blue,dashed] ($(h22)+(-0.4,-0.4)$) -- (4,0);
        \draw[plot_blue,dashed] ($(h22)+(0.4,-0.4)$) -- (7,0);
        \draw[plot_blue,dashed] ($(h22)+(-0.4,0.4)$) -- (4,2);
        \draw[plot_blue,dashed] ($(h22)+(0.4,0.4)$) -- (7,2);
        \fill[white] (4,0) rectangle ++(3,2);
        \fill[plot_blue,opacity=0.05] (4,0) rectangle ++(3,2);
        \draw[color=plot_blue] (4,0) rectangle ++(3,2);
                
        \draw (5.5,1) node[bigneuron] (n) {$n$};
        \draw[->] ($(n)-(1.25,-0.75)$) -- (n) node[near start,anchor=south] {\footnotesize $w_1$};
        \draw[->] ($(n)-(1.25,0)$) -- (n) node[near start,anchor=south] {\footnotesize $w_2$};
        \draw[->] ($(n)-(1.25,0.75)$) -- (n) node[near start,anchor=south] {\footnotesize $w_3$};
        
        \draw[->] (n) -- ++(1.25,0) node[near end,anchor=south] {\footnotesize $o_n$};
                
    \end{tikzpicture}
    \caption{An example of a neural network with two input features in the input layer, two hidden, fully connected layers with three neurons each, and one neuron in the output layer. Output $o_n$ of neuron $n$ is defined as a weighted sum of outputs $o_{n_i}$ of its preceding neurons $n_i$ transformed by the activation function $\sigma$, that is, $ o_n = \sigma(\sum w_i o_{n_i})$, for $i \in \lbrace 1, 2, 3 \rbrace$.}
    \label{fig:neural_network}
\end{figure}

\noindent A \emph{Neural Network} (NN) consists of several layers, with each of these layers containing a different number of neurons. Figure \ref{fig:neural_network} provides an example of a neural network with two hidden layers, and the precise architecture of the network is parametric to the problem the network is addressing. 
In the first, input, layer, every neuron represents one value from the input vector. The values of neurons in the later layers are calculated as a weighted sum of their predecessors transformed by an activation function, and these weights are learned during the training phase. The last, output, layer represents the inferred value, which is determined by the weights, activation functions (which define the output of a node given its input), and the vector of features provided to the network.
Since the activation function does not need to be linear to its arguments, neural networks are able to identify complex patterns in highly non-linear data sets, as it is the case for well logs. \\ \\
The \emph{Gradient Tree Boosting} model (GB) uses decision trees as the building blocks. 
Every branching node in a decision tree corresponds to an if-else condition on one of the features and splits the training samples into two parts, and the condition is chosen so that each split reduces the overall error. Leaves of the tree contain the average target value of the samples, corresponding to each specific leaf and serve as the predicted value. For every new sample that has to be evaluated, the tree is traversed from top to bottom, and left or right sub-trees are taken depending on the features of the sample.

\begin{figure}[hbt]
    \centering
    \begin{tikzpicture}[
        circ/.style={circle,draw,inner sep=0,minimum size=5pt}
    ]
        // First tree
        \draw (0,10) node[circ] (t11) {};
        \draw (-0.25,9.5) node[circ] (t12) {};
        \draw (0.25,9.5) node[circ] (t13) {};
        \draw (0,9) node[circ] (t14) {};
        \draw (-0.5,9) node[circ] (t15) {};
        
        \draw (t11) -- (t12);
        \draw (t11) -- (t13);
        \draw (t12) -- (t14);
        \draw (t12) -- (t15);
        
        \draw (0,10.5) node {$T(\vec{x})$};
        
        // Second tree
        \draw (2,10) node[circ] (t21) {};
        \draw (1.75,9.5) node[circ] (t22) {};
        \draw (2.25,9.5) node[circ] (t23) {};
        \draw (2,9) node[circ] (t24) {};
        \draw (2.5,9) node[circ] (t25) {};
        
        \draw (t21) -- (t22);
        \draw (t21) -- (t23);
        \draw (t23) -- (t24);
        \draw (t23) -- (t25);
        
        \draw (2,10.5) node {$T_{e_1}(\vec{x})$};
        
        // Third tree
        \draw (4,10) node[circ] (t31) {};
        \draw (3.75,9.5) node[circ] (t32) {};
        \draw (4.25,9.5) node[circ] (t33) {};
        
        \draw (t31) -- (t32);
        \draw (t31) -- (t33);
        
        \draw (4,10.5) node {$T_{e_2}(\vec{x})$};
        
        // Signs
        \draw (1,9.5) node {+};
        \draw (3,9.5) node {+};
        \draw (5,9.5) node {=};
        
        // Explanations
        \draw (6,10.5) node {Prediction};
        
        \draw (0,8) node[text width=2cm,align=center] {\footnotesize Prediction with error $e$};
        \draw (2,8) node[text width=2cm,align=center] {\footnotesize Reduces error by $e_1$};
        \draw (4,8) node[text width=2cm,align=center] {\footnotesize Reduces error by $e_2$};
        \draw (6,8) node[text width=2cm,align=center] {\footnotesize Overall error $e-(e_1+e_2)$};
        
        \draw (5,9.5) node[anchor=west,text width=2cm,align=center] {\footnotesize%
        $\begin{aligned}%
          T(\vec{x})&\\
        + T_{e_1}(\vec{x})&\\
        + T_{e_2}(\vec{x})&
        \end{aligned}$
        };
    \end{tikzpicture}
    \caption{The first tree ($T$) is a decision tree providing a rough estimation of the value to be predicted. The second tree ($T_{e_1}$) predicts and corrects the error of $T$, leading to a reduced overall error. However, $T_{e_1}$ also has an error, which can be predicted and further reduced by constructing $T_{e_2}$. This process can be repeated until the desired accuracy is reached.}
    \label{fig:gradient_boosting}
\end{figure}
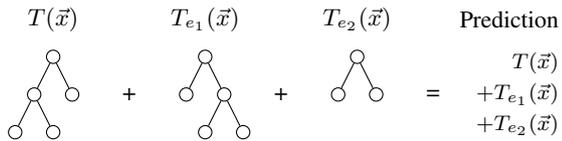

The maximum depth of the tree (i.e. the number of the if-else split conditions), as well as the minimum number of samples that need to be present in a node to allow further branching restricts the size of the tree and precision of the predictions. These are provided as configuration parameters, whereas the splitting conditions are derived during the training phase from the data.

These regression decision trees have severely limited accuracy, as they can effectively predict only the set of values represented by their leaves. Gradient Tree Boosting overcomes this limitation by constructing an ensemble of trees and this can be seen in Figure~\ref{fig:gradient_boosting}, where every consecutive tree corrects the error of the previous tree, refining the prediction until the error converges or the model starts to overfit. \\ \\
\noindent We have also fitted a linear regression model to serve as a baseline to compare the other models against. \emph{Linear Regression} (LR) is a statistical model which uses a linear function to approximate the relationship between the input (explanatory) variables $x_1, \ldots, x_p$ and the target (response) variable~$y$. The general formula is 
$y = \beta_0 + \beta_1 x_{1} + \ldots + \beta_p x_{p} + \epsilon $ , and the model is fitted to the observed (training) data such that the learned coefficients $\beta_0, \ldots, \beta_p$ minimise the error $\epsilon$ across the training data according to some cost function. This is usually the mean squared error, but other cost functions are permissible too. As such, linear regression models are computationally simple but offer only limited accuracy when approximating complex relationships.\\ \\

\subsection{Related Work}
The interpretation and modelling used in hydrocarbon discovery and extraction process relies heavily on the engineer's experience and knowledge of additional, non-logged information. These include facets such as drilling reports and images of the material retrieved from the borehole, and as such are not a prime target for automated computational approaches. However a hypothesis is that the \emph{experience} of petrophysicists might be implicitly contained in a set of well logs of a sufficient size. Several authors have explored the application of ML techniques to related problems, and in this section we present an overview of the most relevant work. 

Liu and Sacchi~\cite{LiuSacchiSVM} explore the use of seismic measurements to propagate information about physical rock properties from given (known) well logs to proposed (incomplete) ones. While their results are promising, the seismic measurements are an additional input which, in the current petrophysical workflows, is used to validate rock physics models derived from the well logs rather than as an input. Using the seismic measurements at an earlier stage of the interpretation workflow would prevent such a validation and therefore we chose to omit the seismic measurements in our work.

Holmes et al.~\cite{HOLMES2003} address the problem of missing data within well logs by ML techniques based solely on the available rock physics measurements. The paper enumerates several pitfalls and provide a guideline on how to fill the gaps. Primarily they state that the common way of applying neural networks does not work for well logs, training the model on available data and afterwards predict the missing measurements, since the geology changes over intervals and therefore might lead to unreliable results. 
Instead, they suggest one selects intervals with similar geological properties for training and to preprocess the data by hand as well as established algorithms to ensure that the model does not learn wrong inferences. Furthermore, the authors recommend using deterministic petrophysical and stochastic modeling along with neural networks to enable experts to choose the most realistic predictions among the three. Although this approach might be more accurate and reliable, it has the disadvantage of requiring a human \emph{in the loop} and therefore does not scale to a large number of well logs and also introduces a bias.

Lopes and Jorge \cite{LOPES2018} trained different ML models on a set of 8 well logs to predict the missing values, disregarding the potential problems caused by geologically inconsistent behaviours in the logged data. In their paper they carry out a detailed descriptive and exploratory data analysis of the gaps and afterwards evaluate Gradient Tree Boosting, Random Forests, Artificial Networks and linear models on the prediction of \nphi{}. We extend their work by using a larger data set and evaluate the models on four different prediction targets. We also investigate how the models' accuracy is influenced by the amount and geographical proximity of the training data. 

Zhang et al.~\cite{Zhang2018} presented approach to filling in the missing measurements using Recurrent Neural Networks, specifically, a cascaded LSTM. The authors claim that the obtained results are more accurate than those predicted by a traditional Neural Network. While we agree that the LSTM can better utilise spatial dependencies within a well log, the presented work reports results from one well log only, and in other work \cite{swoop_cug} it was found that it is  far more challenging to train a model which is general enough to provide good predictions for a large number of wells. Furthermore, we have observed that some well logs are easier to complete than others, which makes a fair comparison with approaches based on their ability to complete one single well log challenging.

\section{Predictive Model}%
\label{sec:method}
Given a raw well log, we automatically complete it by filling the gaps with values predicted  from the other available measurements. In this section we present a tool that uses the most accurate approach to predict the values. To explain this approach, we first describe the available data, then discuss the problem in more detail, ultimately proposing a robust set of evaluation metrics and exploration of the model.

\subsection{Data and Exploratory Analysis}

This  work was focused on the \nlog{}\footnote{https://www.nlog.nl/en} data set, a publicly available set of well logs from the Netherlands and the Dutch sector of the North Sea continental shelf. The provided well logs contain only raw measurements without any human interpretation, as this additional information is proprietary and of commercial value to companies who provide surveying services. 

The well logs are plain text files formatted according to \las{} standard~\cite{LAS}, and the header of the log file contains general information about the log. This includes properties such as the name of the well and its geographical coordinates, the depth at which the borehole begins and ends, lists of properties measured and units in which they are recorded\footnote{This is rather important as some tools use meters and other feet as basic units and so the logs need to be checked and values converted accordingly.}, and which value represents missing measurements (nulls). The remainder of the file is organised where each row corresponds to a specific depth in the well, and every tab-separated column records a value of one measured property. The measurements are recorded at regular depth increments, with the step size explicitly stated in the header of the \las{} file, but typically between 10 and 15cm.

\begin{wrapfigure}{r}{0.25\textwidth}
    \includegraphics[width=\linewidth]{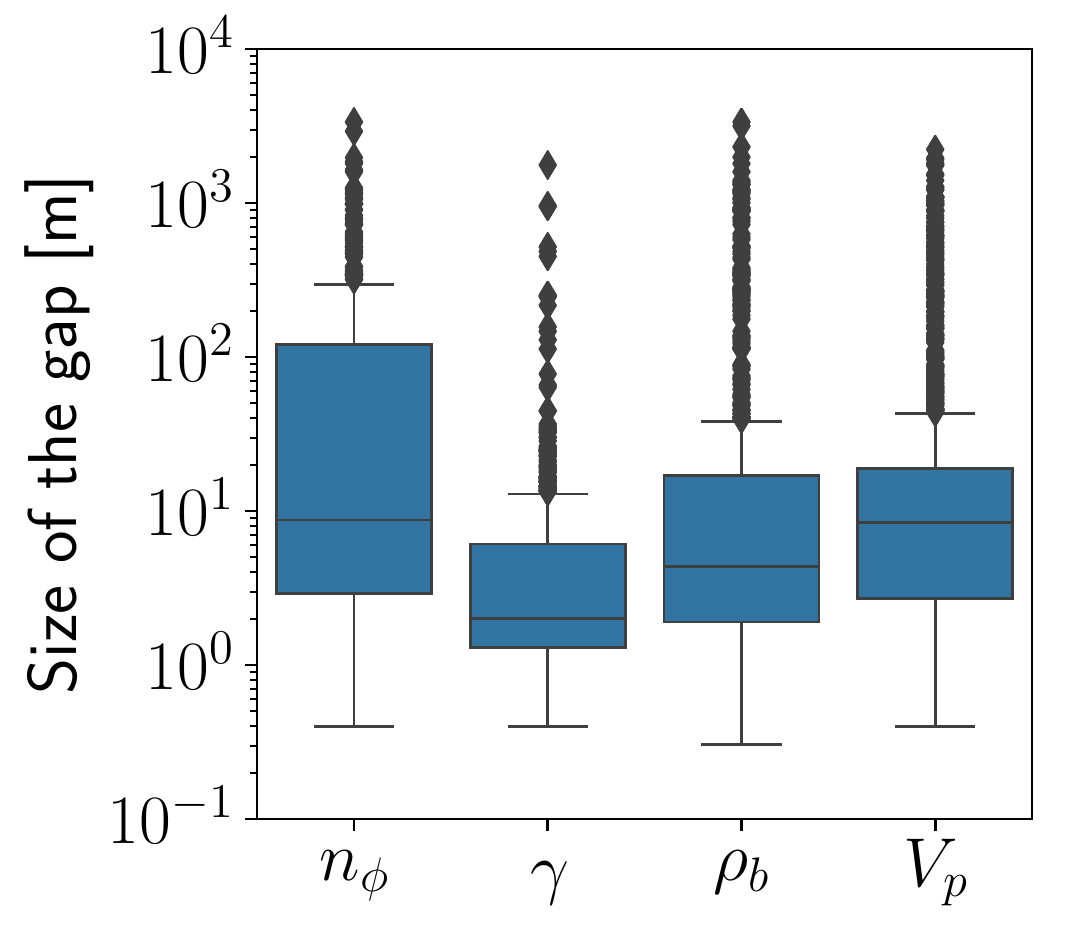}
    \caption{Box plot illustrating the average sizes of the gaps in the \nlog{} data set for all features.}
    \label{fig:gap_sizes}
\end{wrapfigure}
The properties recorded across most of the available well logs are \vp{}, \nphi{}, \gr{} and \rhob{}, however, these values are not recorded for every row. These missing features create gaps in the well logs which  reduces the information available and introduces uncertainty about the properties of the surrounding geology. For illustration, the \nlog{} data set contains more than $23~\cdot~10^6$ rows of measurements, but only 26\% of those are complete.

In this work, we define a \emph{gap} as a sequence of missing measurements of any property that spans more than 0.3m. The only exception are sequences of intentionally omitted measurements from the top of the borehole. These omissions are either due to the presence of sea water rather than solid ground, or because the engineers on site judged that the measurements at such a shallow depth would not have significant commercial value. In either case there is no need to predict these intentionally omitted values. 
The gaps in the \nlog{} data set have average sizes ranging from less than \SI{15}{\meter} (for \gr{}) to \SI{170}{\meter} (for \nphi{}) (see Figure~\ref{fig:gap_sizes}), are distributed randomly through the depth of the well and occur independently from each other.

\subsection{Evaluation Criteria}%
\label{sec:artificial_gap_function}

We address the problem of predicting missing values by supervised ML. To build a predictive model, we first selected a set of training data, from which the model derives relationships among input and target variables. A test set is also identified, which enables us to measure the accuracy of the model. For both of these data sets we need to know what target values the model should predict, so we cannot use portions of the well logs which contain real gaps. 

To enable a comparison of predictions for different target properties and different training set-ups, we selected a set of test wells and inserted \emph{artificial gaps} across all four properties at the same depth.
However, in some cases the well logs contained only relatively few samples and splitting these into training and test sets would have resulted in a very limited amount of data for training. 

To maximise the size of the training data set we filter on wells which meet the following criteria: a minimum depth of \SI{1.5}{km},
a maximum gap size of \SI{50}{m} and a ratio between complete and incomplete samples of at least $0.5$.
The maximum gap size ensures that there are no large uncovered intervals within the well and the ratio makes sure that the well contains at least \SI{750}{m} of valid log data. 

After removing wells which do not meet these criteria we are left with 60 valid wells left, containing 1,335,757 samples in total. We selected 50 random wells for the test set and this represents approximately 7\% of the overall number of wells in the \nlog{} data set, but more than 80\% of the usable wells.

\noindent The probability that a value will not be measured at a specific depth is different for each property: \vp{} is omitted most frequently with 42.9\% of rows missing this feature, followed by \gr{} which is absent at 23.4\% of depths. The probabilities the \rhob{} or \nphi{} will be missing are 23.0\% and 10.6\%, respectively. 

We also investigated whether the missing measurements correlate among the logged properties, defining a \emph{gap} as a sequence of consecutive missing measurements in a property log. We represent these gaps as nodes of an undirected graph, inserting an edge if the starting depths and sizes of the gaps corresponding to the end nodes do not differ by more than 10m and 10\% respectively, marking the gaps as \textit{coinciding}. Every connected component of this graph then corresponds to a part of the well log with missing measurements, and the order of the component corresponds to the number of properties affected. 
We discovered that the missing measurements are rarely correlated among different properties. 83\% of the components are of order 1, corresponding to measurements missing only in one property log at a time. Gaps coinciding in two properties occur in 14\% of cases, and the probability of three or all four properties missing are 1\% and 2\%, respectively.
These results demonstrate that there is no obvious correlation between the occurrence of a gap and geological or morphological properties of the rock surrounding the borehole, which would affect the other measurements.

By scanning the data set we observed that rows with 1 measurement missing comprise more than 98\% of the incomplete data, while rows missing more than one measurement represent approximately 1.5\% of the available data. As such, predicting one property by relying on the other three is a realistic proposition and be of benefit across much of the data set. 

\begin{wrapfigure}{l}{0.175\textwidth}
    \centering
    \includegraphics[width=\linewidth]{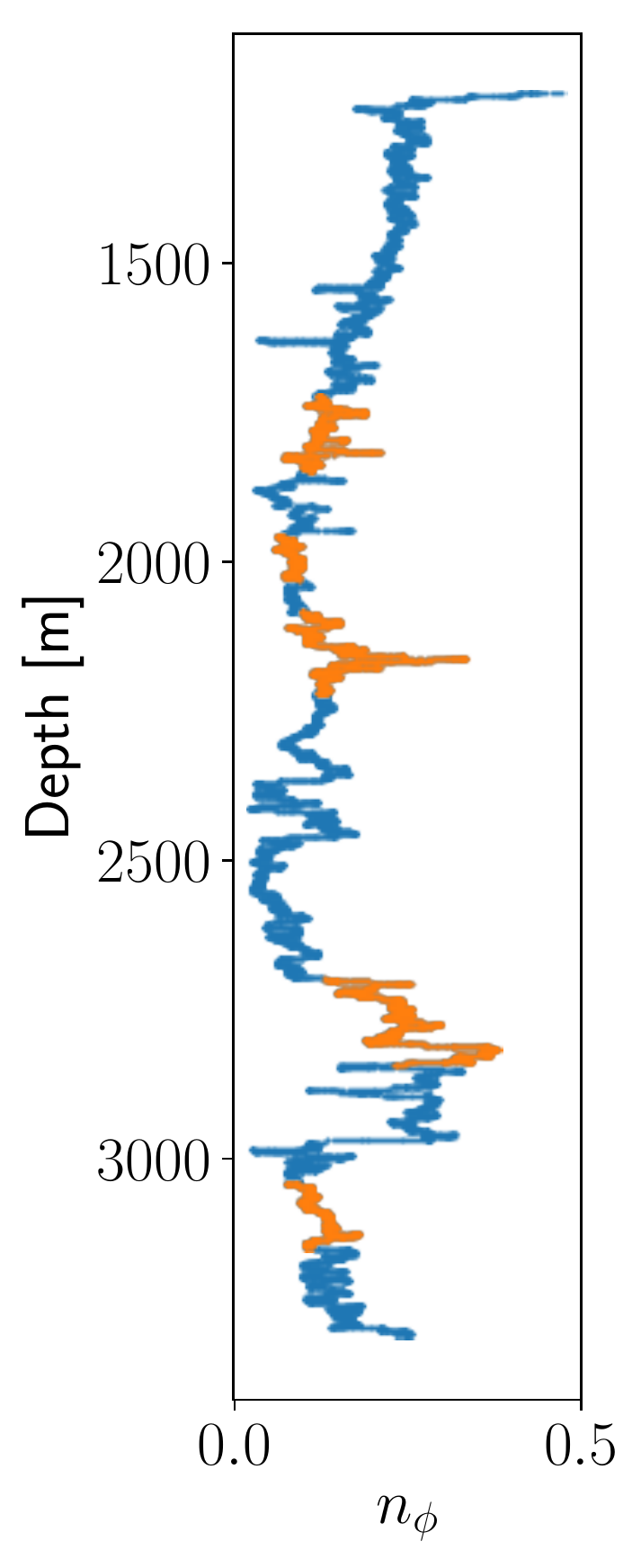}
    \caption{Artificial gaps (orange) in an \nphi{} log with gap size of $\SI[separate-uncertainty=true]{150(50)}{\meter}$ and 2 gaps per km.}
    \label{fig:gaps}
\end{wrapfigure}

\noindent We designed a function that, given statistical information about gaps (mean and standard deviation of the size, and their average number per km of depth) and a well log, creates random gaps throughout the borehole. The sizes and positions of the generated gaps are random but follow the distribution observed in the data set: 1) The size of the gaps correlates negatively with the relative depth they appear at, with gaps at the beginning of the borehole more likely to be larger, and 2) gaps are more likely to occur at the end of the borehole, with a probability increasing linearly from the beginning to the end.

In this work we generate gaps of $\SI[separate-uncertainty=true]{150(50)}{\meter}$, where the average size of an artificial gap was selected based on the average sizes of gaps in the well logs. We have observed that predicting smaller gaps is easier, as there are more similar measurements available in the training data, and so we bias the artificial gaps to be of larger sizes. This means that we are able to guarantee accuracy even for the properties most difficult to predict.


In the \nlog{} data set there are on average 1.6 gaps per km, although this number is highly variable, with some wells containing up to 30 gaps per km. When generating the artificial gaps, we feed into the function the value \emph{two} as the coefficient of the number of gaps per km.
\\ \\
%
The \emph{Mean Squared Error} (MSE) measures the average square of the differences between the predicted and true values, and is the default loss function used for regression problems. As such we adopted the MSE as the loss function in our predictive models, and also as the main quantitative measure when evaluating and comparing the models. 

We find that to interpret MSE correctly, one needs to consider the target's range, as an error of 100 might be acceptable for property with values measured in thousands, but would imply a rather poor prediction for a property with average value in units. As we are considering several different target variables with rather diverse ranges of values, we also report \emph{Mean Absolute Percentage Error} (MAPE), which calculates the relative difference between true and predicted values, allowing us to compare the accuracy of the models for the target properties with different ranges straight away. 

It is important to highlight that both of these metrics provide only an approximation of the model's performance. This is because they report only one average number, and the way this is calculated might hide significant information about the fit of the prediction in the context of the borehole. For instance, a model that consistently under-predicts the magnitude of the measurements, but generally exhibits the right pattern of whether property values are increasing or decreasing, might be preferable to a model that predicts some values very closely, but misses others by a lot without any consistent pattern, effectively predicting noise. Therefore, especially when comparing the accuracy of different models, we plot the predictions and examine these manually. \\ \\
%
%
The test rows, used to evaluate the predictions, cannot be sampled entirely randomly as they belong to one of the test wells selected beforehand and need to be grouped to form gaps similar to the real ones, which is accomplished by generating the artificial gaps. 

We evaluate the predictions across these artificial gaps quantitatively by means of MSE and MAPE. However, the average numbers might hide some bias or other systematic information about the quality of the predictions, and so we also look at the predictions individually in the context of the whole borehole. Such visual inspection introduces also a qualitative evaluation of the predictions.

\subsection{Features}%
\label{sec:features}
The choice of input features to the ML model is crucial and this has a huge impact on the accuracy and validity of the results. Since we want to minimise the effort required by the engineers working with the logs, we mainly utilise properties recorded in the well logs: \nphi{}, \gr{}, \rhob{} and \vp{} and features that can be automatically derived from them, which might provide more contextual information to the models. The input vector consists of:
%
\begin{itemize}
    \item three rock physics properties measurements\footnote{These were normalized by using the \texttt{RobustScaler} of the sklearn python package before feeding them into the model.}, as we have observed that the gaps appear independently of each other,
    \item the average value for the four rock properties in the given well in order to provide well-specific contextual information,
    \item the depth at which the measurements were taken.
\end{itemize}




\subsection{Model Description}
As already noted in \cite{HOLMES2003}, the characteristics of the subsurface are very diverse and can change even over small distances.
In this section we describe an approach in which linear regression, the Gradient Boosting and neural network models are trained only on the well itself to predict missing values. To evaluate this approach we rely on the artificial gap function introduced in Section \ref{sec:artificial_gap_function}, 
which splits the logged data into training and test set by inserting gaps to various depth levels of the well where the ground truth is known, allowing us to evaluate the model against it. 

To implement the Gradient Boosting model, we used the \texttt{GradientBoostingRegressor} of the Python sklearn library\footnote{https://scikit-learn.org/}. We found empirically that the default parameters are already well tuned and fitted to our problem and so they do not have to be adjusted. The model builds up to 100 trees with a learning rate of 0.1, a maximal depth of 3 and a minimal number of 2 samples per split.

In case of the neural network we used the Keras\footnote{https://keras.io/} library with a three dense-layer network consisting of 20, 15 and 1 neuron(s). The Nadam optimiser was used in combination with the MSE as our loss metric, and we trained the network over 200 epochs, but used an early stopping mechanism to prevent the model from overfitting. The \texttt{relu} activation function was used for the first two layers, and a linear activation function for the output layer. Before the features were handed over to the input layer they were scaled, as this is expected to increase the accuracy of neural networks in general. 

Hyperparameters were derived for both models by starting with the default values and then changing these slightly in an iterative process, feeding back depending on the models' reaction. We also explored the use of automated hyperparameter tuning via the hyperopt\footnote{http://hyperopt.github.io/hyperopt/} library, performing a guided search through the space of hyperparameters. Although we considered different network architectures, optimisers, activation functions and learning rates in our search, this automated approach did not lead to any significant improvement in accuracy. However, we have observed that the GB accuracy is considerably more stable than the one of NN, which is highly dependant on the random initialisation of the network's weights.

\begin{table}
\centering
\begin{tabular}{l l | cccc}
\toprule
 &	&	\nphi{}	& \gr{}	& \vp{}	& \rhob{} \\
\midrule
\multirow{ 3}{*}{MSE}  
    & NN & $157 \cdot 10^{-5}$ & 227~ & $41 \cdot 10^3$ & $1574 \cdot 10^{-5}$ \\
	&GB & ~$135 \cdot 10^{-5}$  & 206~ & $31 \cdot 10^3$ & $804 \cdot 10^{-5}$ \\
	& LR & $306 \cdot 10^{-5}$ & 335~ & $65 \cdot 10^3$ & $1250 \cdot 10^{-5}$ \\
\midrule	 
\multirow{ 3}{*}{MAPE} 
    & NN & 45.37~~~ & 12.51 & 3.75 & 2.92~	\\
	& GB & 50.44~~~ & 13.22 & 3.29 & 2.27~	\\
	& LR & 140.64~~~ & 20.36 & 4.96 & 2.97~	\\
\bottomrule
\end{tabular}
\caption{Accuracy of all targets predicted by GB, NN and LR models, re-trained for every test well and averaged afterwards.}
\label{tab:results}
\end{table}

\subsection{Evaluation}
Table~\ref{tab:results} illustrates the results of evaluating our two approaches by fitting the models to wells sampled for testing, and compares them against linear regression model. The fitting required training the different models for each of the four target properties and for every test well, using the artificial gaps as testing data and the remaining complete rows as training data. The MSE and MAPE metrics reported in Table~\ref{tab:results} for each configuration illustrate the average performance for each model across all the wells. Whilst both GB and NN predictions are within acceptable limits, it can be seen that the GB approach consistently generates more accurate predictions than the NN, and the LR does not outperform the other models.

%
\begin{figure}[thb]
    \centering
    \includegraphics[width=.9\linewidth]{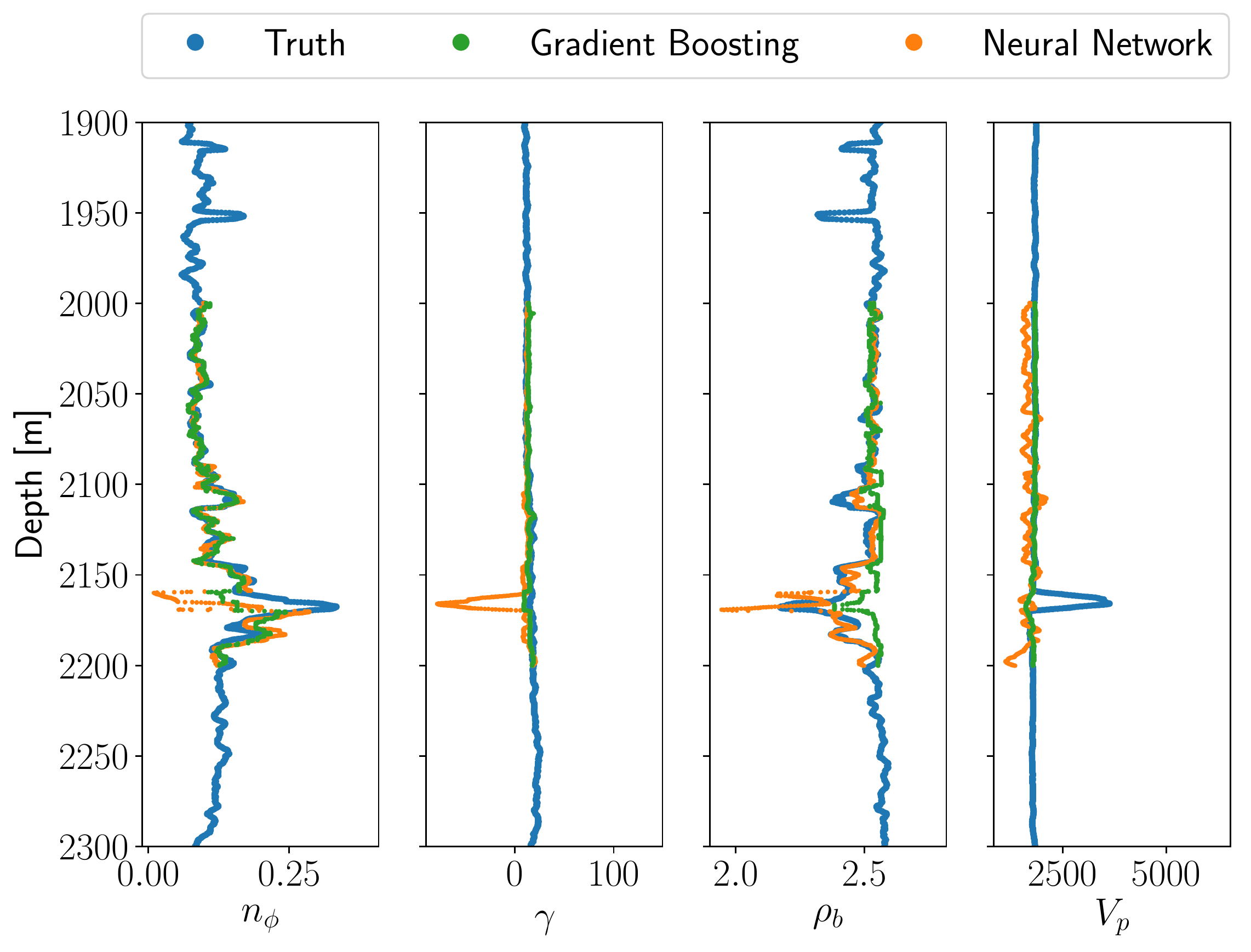}
    \caption{Well L17-02 with completed artificial gaps by both models.}
    \label{fig:predictions}
\end{figure}

\noindent During our experiments we also found that the error of \nphi{} depends on the choice of the test well, as there are a few outlier wells with errors higher than 1000\%, which then drive up the mean error.

\begin{wrapfigure}{r}{0.45\linewidth}
    \includegraphics[width=\linewidth]{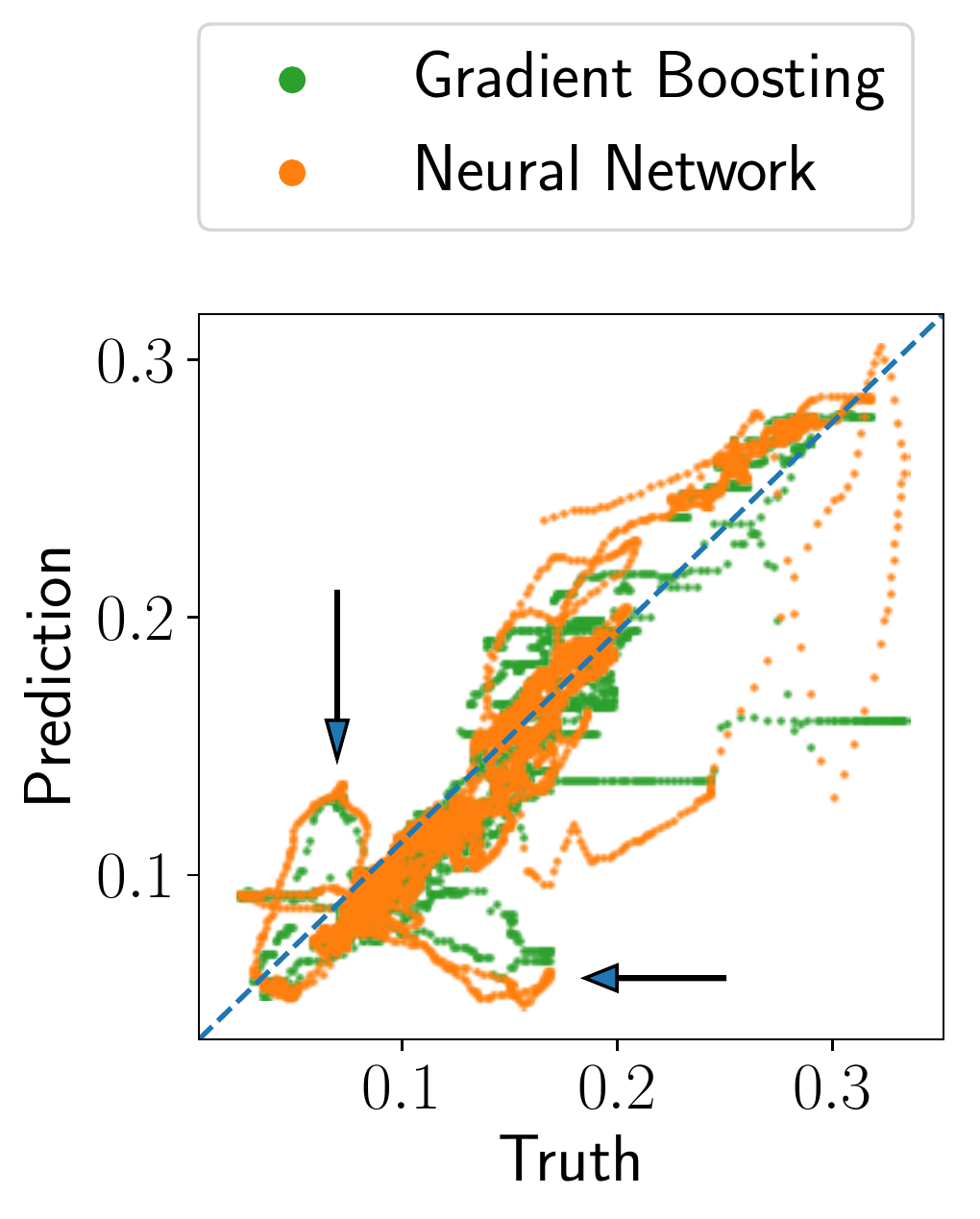}
    \caption{Predicted measurements for \nphi{} on the $y$-axis versus the ground truth on the $x$-axis for well L17-02.  Arrows indicate wrong patterns learned by both models.}
    \label{fig:truth_vs_pred}
\end{wrapfigure}
For \gr{}, \vp{} and \rhob{}, the percentage errors are low and well within acceptable range. In contrast, \nphi{} demonstrates a relatively large percentage error, but that was to be expected because \nphi{} is not strongly linked to the other geological properties. 

Figure~\ref{fig:predictions} illustrates the ground truth of well L17-02, including artificial gaps and their predicted values for both models. The predictions of the gaps in the \rhob{} curve are an example of the quantitative errors being misleading, where numerically both the GB and the NN models perform with nearly identical accuracy, missing the target by 2\% on average. However, one can easily see that the NN predictions (orange) have similar magnitude in peaks to the true measurements (blue), while the prediction of the GB model (green) is more conservative, with the values centered around the true values.

To identify any systematic bias of the two models, we plot the truth against the predicted values for well L17-02 in Figure~\ref{fig:truth_vs_pred}. Ideally, the points should be clustered around the diagonal line, and in Figure~\ref{fig:truth_vs_pred} we include only \nphi{} as this is sufficient to illustrate two important findings. Firstly there does not seem to be a general systematic bias, as the predictions are located close to the ideal diagonal line. 
Secondly, sometimes both models learn identical relationships between the features, leading to a similar pattern in Figure~\ref{fig:truth_vs_pred}. This indicates that there is a strong correlation in the data, which both models identify, but which does not correspond to a real relationship between the features. 




\section{Extending the Data set}%
\label{sec:alternative_approaches}

The previous section reported results of ML models trained on a single well in order to predict values missing in that specific well. While this approach is acceptable for some of the properties, such as \vp{} or \rhob{}, it is far from perfect. In this section we demonstrate the impact of larger data sets for training and considering the geographical proximity of wells.


%

\subsection{Global Training}
\label{sec:global_model}
%
Extending the work of Lopes and Jorge~\cite{LOPES2018}, we trained the two ML models on all the available data except the gaps that were generated for evaluation purposes in Section~\ref{sec:method}. We still needed to train one model for each property, but, different from before, this model will now predict all the gaps across the whole test set at once, without requiring a re-train for each well. The hypothesis was that the larger training set would better represent the variety of possible geologies surrounding the wells.

\begin{table}
\centering
\begin{tabular}{l l | cccc}
\toprule
 &	&	\nphi{}	& \gr{}	& \vp{}	& \rhob{} \\
\midrule
\multirow{ 2}{*}{MSE}  
    & NN & $2313 \cdot 10^{-5}$ & 1769~~~~~ & $1019 \cdot 10^3$ & $7831 \cdot 10^{-5}$ \\
	&GB & ~~$311 \cdot 10^{-5}$  & 707~~~~~ & $108 \cdot 10^3$ & $1451 \cdot 10^{-5}$ \\
\midrule	 
\multirow{ 2}{*}{MAPE} 
    & NN & 699.21~~~~~ & 81.96 & 19.57 & 8.72~	\\
	& GB & 588.57~~~~~ & 95.03 & 16.41 & 7.94~	\\
\bottomrule
\end{tabular}
\caption{Accuracy of all targets predicted by GB and NN models, both trained on the full training data set.}
\label{tab:results_global}
\end{table}





\noindent Table~\ref{tab:results_global} illustrates the prediction error of the trained models, where the MSE is generally one order of magnitude higher (worse) than when predicting the gaps by models trained for each well individually. The MAPE is also proportionately higher than it was when trained on a single well as in Table \ref{tab:results}. The GB model is again more accurate than the NN approach.

\begin{figure}[htb]
    \centering
    \includegraphics[width=.9\linewidth]{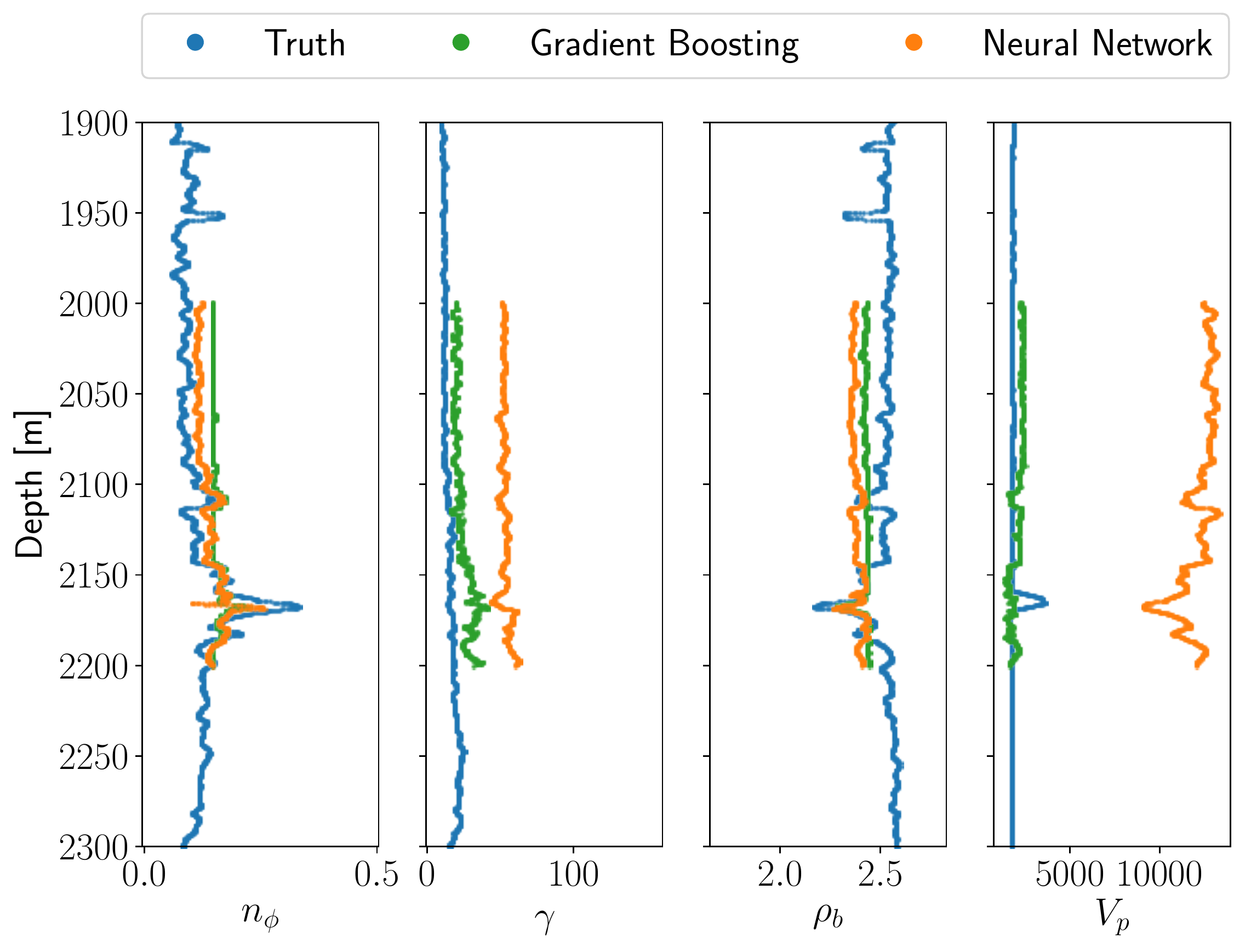}
    \caption{Plot of well L17-02 including predictions of both models after they were trained on the whole training set.}
    \label{fig:predictions_global}
\end{figure}


\noindent Figure~\ref{fig:predictions_global} presents the predictions for well L17-02, the same well as in Section~\ref{sec:method} but now gaps are predicted by models that were trained on the entire data set (except the previously generated gaps). Several facets can be observed from these results: 
\begin{itemize}
    \item Both models are now less sensitive to the peaks in the input curves when predicting \gr{}, the false peak is much less pronounced. 
    \item The GB model's predictions have slightly greater amplitude, which is more realistic of real-world values.
    \item All the predictions have a significant offset from the true values.
\end{itemize} 

\noindent While the greater variability of the predictions and the general shape of the predicted curves is more representative of the behaviour of real measurements, the offset presents a serious problem. Note that there is the average value of each property for every well among the input features, so we would expect the models to be able to align the predictions closer to the true values. Moreover, this offset does not happen consistently in all predictions, not even in all predictions in one gap, so there is no obvious way to align the predictions with the magnitude of the property e.g. by post-processing. 

%
\begin{figure}[htb]
    \centering
    \begin{tabular}{c c}
        \includegraphics[width=0.37\linewidth]{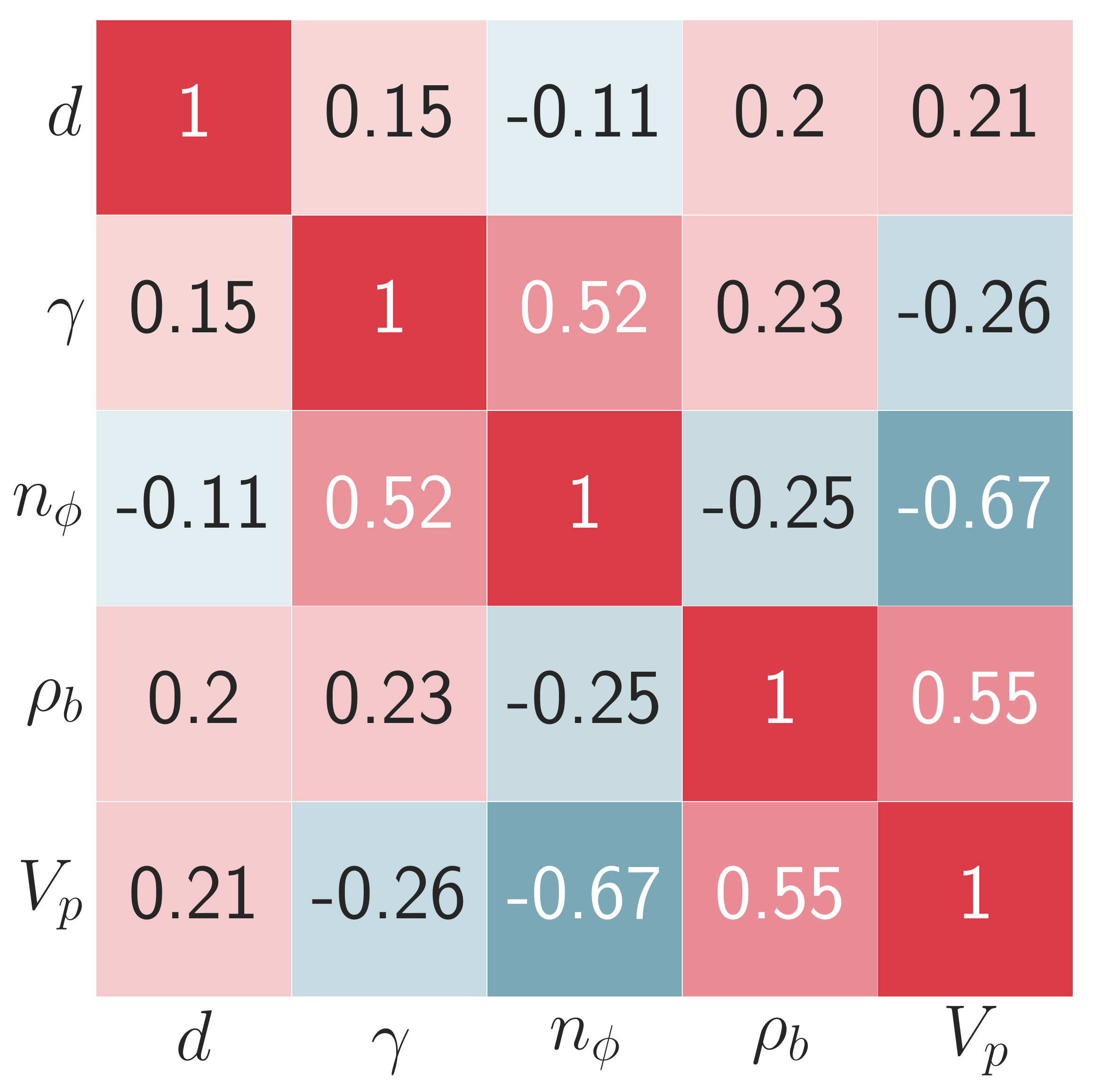} &
        \includegraphics[width=0.37\linewidth]{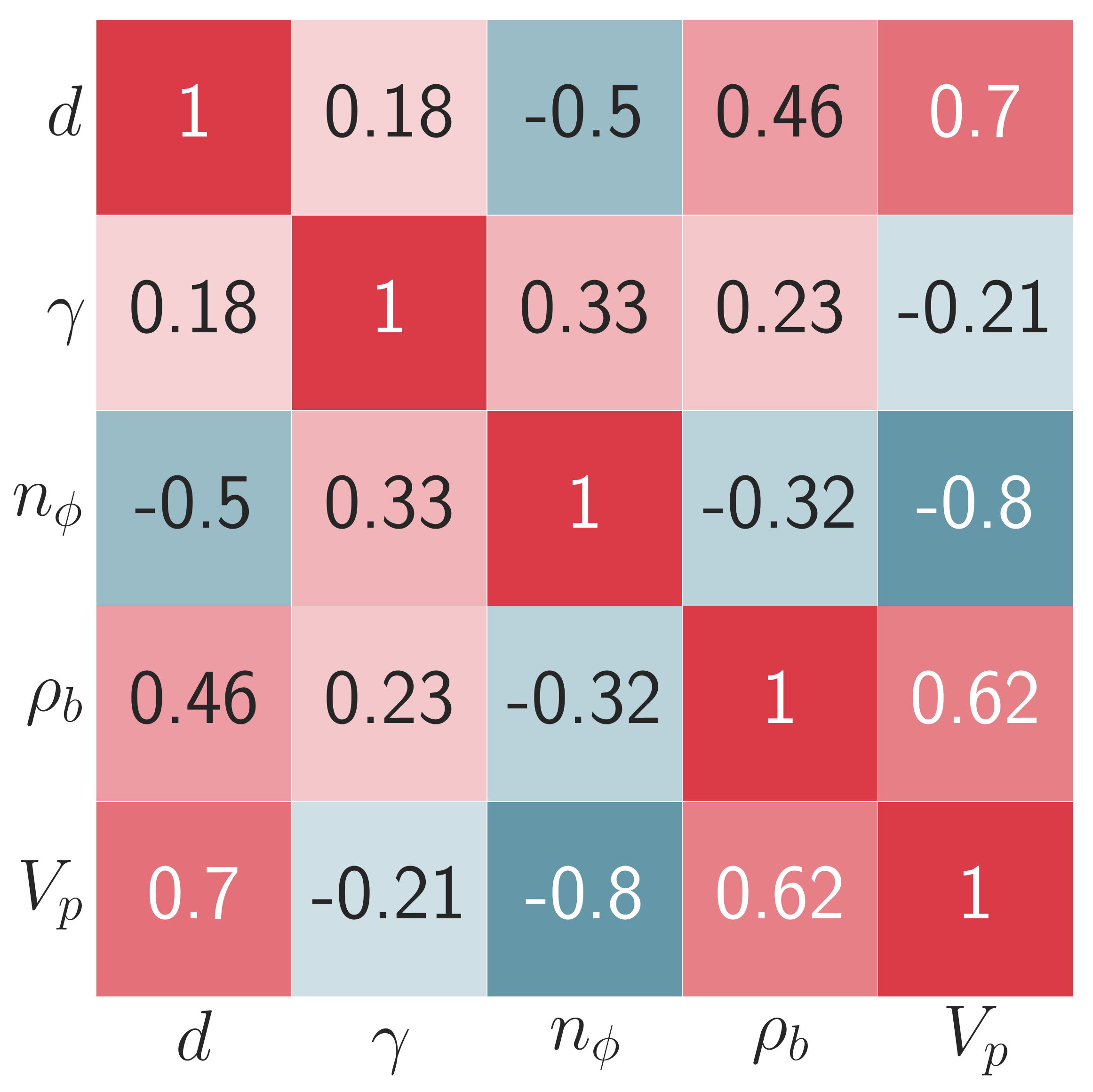} \\
        (a) & (b) \\
    \end{tabular}
    \caption{Correlations of the properties: (a) for every well, averaged, and (b) over the whole data set.}
    \label{fig:correlations}
\end{figure}
\subsection{Local Training}
Training on all wells at once does not improve the quality of results. Whilst there appears to be stronger correlations among the features, many of these are most likely random: Figure~\ref{fig:correlations} contains a correlation heatmap calculated for all properties across the whole data set (on the right, b)), contrasted against correlations found in every well, which are captured by the heatmap on the left, (a)) as average correlations on per-well basis.
Note for example that \nphi{} and \gr{} are not correlated in a statistically significant way when looking at each well separately.
Instead, we attempt to identify number between \emph{one} and \emph{all other} wells which would allow the model to learn better inferences, but not coincidental similarities. The idea is that wells in close geographical proximity are more likely to have undergone similar geological processes and hence might exhibit similar geophysical relationships. 


We tested this hypothesis by taking five random wells from the test set and training GB and NN models for each of these which are \emph{baseline} models. Note that these models coincide with the models studied in the previous section. 

\begin{figure}
    \centering
    \includegraphics[width=0.95\linewidth]{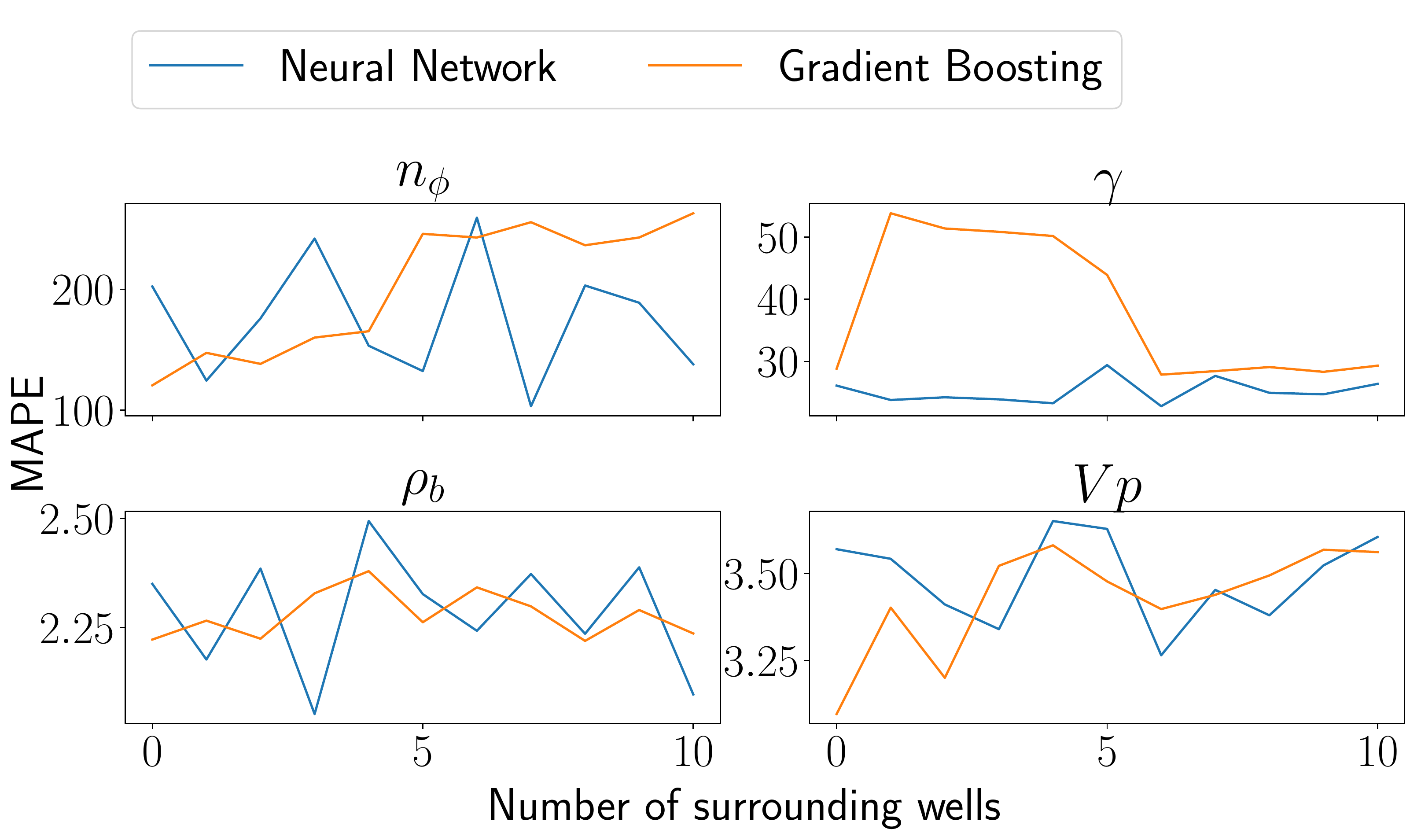}
    \caption{
    This plot shows the evolution of the MAPE for each property when surrounded wells were added as further training data. When the number of surrounding wells is 0, only the well itself was used during the training process and its artificial gaps for evaluation.
    }
    \label{fig:close_wells_mape}
\end{figure}

Using the distance to each starting well, we then picked the next closest well and added it to the training data of the baseline model, re-fitting the model and predicting the starting well. We do not leave out parts of the newly added well since we are only interested in the artificial gap of the starting well. Consequently, the test samples are never extended when adding further wells.
We kept adding wells and tracking the errors after each iteration. This process was repeated until we reached the 10 nearest wells, and the evolution of the models' MAPEs is depicted in Figure~\ref{fig:close_wells_mape}.

\noindent From Figure~\ref{fig:close_wells_mape} it can be seen that the information gain of training on additional well log is not a function of the well's geographical distance, and the geological differences observed even on a relatively small geographical scale confuse any pattern the models could learn. Adding more data becomes counter-productive rather fast.

\section{Conclusions and Future Work}%
\label{sec:conclusion}
The petroleum industry has vast amounts of recorded well log data at its disposal, but analysing them is a very time-consuming process, and the condition of the unprocessed data is too poor to be utilised directly. In this paper we demonstrated how ML can exploit such data to build predictive models that can improve the condition of the well logs. 
We show that the differences in geology are significant, even on a small geographic scale, and that the characteristics of the borehole strongly depend on its location. Consequently, the location of the training data affects the models' accuracy and adding more training data does not improve the model in terms of quantitative errors. 

Having identified the optimal amount of training data, we compared a Gradient Tree Boosting and a neural network model on the prediction of missing well log data. Though the Gradient Boosting model does not always outperform the neural network with regard to the error metrics, its performance is more stable -- neither the choice of a particular well log nor the hyperparameters have a particularly severe impact on the level of accuracy, while the prediction accuracy of the neural networks changes in every run.
%
We have demonstrated that, for most of the properties, the GB models can accurately predict gaps with over-average sizes.
We have observed that the common quantitative error metrics (MSE and MAPE) often cannot capture the quality of the predictions and therefore might be misleading, and we addressed the problem by plotting the predictions and examining them manually.

Whilst the models themselves, especially the NN, are rather simple, it is our view that to improve the accuracy it is not the models that need to be re-designed, but instead more sophisticated metrics are required, which will guide the training of the models and assess the models' accuracy better, and improved input data. This problem of the variability and ambiguity of the data could be addressed by adding more explanatory features, or by having a petrophysicist assemble a representative set of training samples.

\ack 
We would like to thank Dr Lucy MacGregor for the consultation and insights on the geophysical aspects of this work, and to the reviewers for their constructive comments. We would also like to thank EPCC\footnote{https://www.epcc.ed.ac.uk} for funding this work and providing the computational resources used throughout.

\bibliography{ecai}

\end{document}